\documentclass[10pt,letterpaper]{article}

\usepackage{cogsci}

\cogscifinalcopy 

\usepackage{graphicx}
\usepackage[
  style=apa,
  natbib=true,
  annotation=false,
]{biblatex}
\usepackage{float} 

\title{A Rational Account of Categorization Based on Information Theory}

\author{
  {\large\bfseries Christopher J. MacLellan (cmaclell@gatech.edu)$^1$, Karthik Singaravadivelan$^1$,}\\
  {\large\bfseries Xin Lian$^2$, Zekun Wang$^1$, and Pat Langley$^{3,4}$} \\
  {\normalsize\normalfont
    $^1$School of Interactive Computing, Georgia Institute of Technology \\
    $^2$Department of Computer Science, Northwestern University \\
    $^3$Georgia Tech Research Institute, Georgia Institute of Technology \\
    $^4$Institute for the Study of Learning and Expertise
  }
}

\begin{document}

\maketitle

\begin{abstract}
We present a new theory of categorization based on an information-theoretic rational analysis.
To evaluate this theory, we investigate how well it can account for key findings from classic categorization experiments conducted by \citet{hayesroth1977role,medin1978context,smith1998prototypes}.
We find that it explains the human categorization behavior as well as (or better) than the independent cue and context models \citep{medin1978context}, the rational model of categorization \citep{anderson1991human}, and a hierarchical Dirichlet process model \citep{griffiths2007unifying}.

\end{abstract}

\section{Introduction}
\noindent The underlying premise of a rational analysis is that humans have evolved cognitive mechanisms that maximize utility given the structure of their environment and their operating constraints \citep{anderson1991human,howes2009rational}.
Thus, if we can devise artificial mechanisms that maximize a given human objective under the relevant environment and operating constraints, then it should have a close correspondence to the adapted human cognitive mechanisms---both should converge to isomorphic mechanisms.

To apply this approach to human categorization requires making assumptions about (1) the objective people maximize during learning and performance, (2) the key structural aspects of the environment, and (3) the relevant operating constraints.
\citet{anderson1991human} proposes answers to all three.
Specifically, he assumes that people learn categories that minimize the mean-squared feature predictive error within a Bayesian inference scheme, and that the environment is structured such that objects can be partitioned into disjoint categories that have independent, probabilistic features within category.
Lastly, he assumes people are constrained to maintaining a single hypothesis about category structure (they do not maintain all possible category structure hypotheses) and that this hypothesis is incrementally updated after each new experience.

The current work builds on \citet{anderson1991human}'s assumptions, but explores two changes.
First, we propose a new objective based on information theory; mainly, that people learn categories that have maximal mutual information with the features of their experiences.
This new objective is based on \citeauthor{corter1992explaining}'s \citeyearpar{corter1992explaining} {\it category utility hypothesis}, which states that a category has utility to the extent that it improves one's ability to predict the features of an instance given knowledge of its category label.
Second, we assume that the structure of categories in the environment is taxonomic, being organized hierarchically rather than in a flat partitioning.
This assumption was suggested by \citet{fisher1990structure} and later explored by \citet{anderson1991human}.

\section{Instantiating the Proposed Theory}

The next step in our rational analysis is to devise artificial mechanisms that maximize this objective within the proposed environment and processing constraints.
We instantiate our theory within \citeauthor{fisher1987knowledge}'s \citeyearpar{fisher1987knowledge} Cobweb system. 

Cobweb employs two distinct representations.
First, instances are described using attribute-value lists (e.g., \{``color'': ``green'', ``shape'': ``circle'', ``size'': 2.4\}).
Second, concepts, which are organized in a hierarchical taxonomy, store probability distributions over the attribute values of training instances associated with them. Concepts maintain categorical distributions for discrete features and normal distributions for continuous features. The system has a single free parameter $\alpha$, which is a smoothing parameter representing uniform pseudocounts for every categorical attribute value.

To predict missing features of an example, Cobweb sorts it into its hierarchy and uses the concepts it activates to generate predictions.
Unlike \citeauthor{fisher1987knowledge}'s original performance mechanism, our variant uses best-first search to expand the nodes with the highest pointwise mutual information with the example.
The pointwise mutual information \citep{church1990word,fano1961transmission} between a concept $c$ and an example $x$ is computed as $PMI(c, x) = \log \frac{p(x,c)}{p(x)p(c)} = \log \frac{p(x|c)}{p(x)} = \log p(x|c) - \log p(x)$.
When deciding which node to expand, the $p(x)$ term can be dropped as it is a constant that appears across all PMI scores; thus, this approach reduces to expanding the nodes with the highest $\log p(x|c)$.
This measure, which quantifies the amount of information the concept label gives about the example's features, directly operationalizes \citeauthor{corter1992explaining}'s category utility hypothesis.
Our system keeps expanding concept nodes up to a user-specified {\it max\_nodes} value (we set this parameter to 100 for all our following simulations) or until there are no more concepts to expand.
Finally, the system predicts the target missing features as a combination of the predictions from all activated nodes weighted by each node's pointwise mutual information score.
Note, we normalized the pointwise mutual information scores, so that the weights across expanded concepts sum to 1.
This inference approach is analogous to an unsupervised version of the predictive equation used by \citet{anderson1991human} that assumes a uniform prior over each activated concept.

To update its taxonomy, Cobweb sorts the examples into its hierarchy in a greedy fashion.
At each branch it considers four possible update operations: (1) {\it inserting} the example into one of the children, (2) creating a {\it new} node to represent the example, (3) {\it merging} the two concepts most similar to the example and inserting into the new merged concept (the original concepts become the children of the new concept), and {\it splitting} the most similar concept and promoting its children.
Our variant of Cobweb chooses the operation that maximizes the future expected pointwise mutual information across the children and all possible examples, which is equivalent to maximizing their mutual information: $\mathbb{E}_{x,c}[PMI(x,c)] = \sum_c \sum_{x} p(x,c) \log \frac{p(x,c)}{p(x)p(c)} = I(X,C)$.  
Similar to \citet{anderson1991human}, we assume that features are statistically independent within each category, which makes it possible to compute this value in a closed form for both categorical and normal feature distributions.
When an example is incorporated into a concept, we incrementally update its feature probability distributions to reflect the new example.

Unlike Anderson's approach, which assigns examples to the categories most likely to have generated them even if this lowers the likelihood scores for previously assigned examples, our approach maximizes the likelihood over all examples, both current and prior. 
The two approaches converge to similar assignments as the amount of data grows, but differ  during early learning and when concepts are based on little data. 
Given the space limitations, we refer the reader to more detailed descriptions Cobweb variants similar to ours by \citet{lian2024cobweb,barari2024cobweb4v}.
Our systems main difference is its use of pointwise mutual information rather than collocation. 

\section{Preliminary Tests of the Theory}

\subsection{Central Tendencies of Categories}

As a first test, we applied our Cobweb variant to simulate the data reported by \citet{hayesroth1977role}.
Their experiment, which investigated human recognition and classification of {\it old} and {\it new} examples, showcases the central tendencies of human categorization---that classification accuracy increases the closer an example is to a category's prototype.

\subsubsection{Human Study Design}
In this study, 108 participants reviewed flashcards showing records of fictional individuals that were described by five features (last name, age, education, marital status, and hobby) and were asked to classify them into one of three classes: {\bf Club 1}, {\bf Club 2}, or {\bf Neither}.
The last name and hobby were distractor features; every card featured a different last name and one of three randomly assigned hobbies.
The remaining features, which determined the class of the example, could take on one of four values.\footnote{See Table 1 of \citep{hayesroth1977role} for the stimuli.}

During training, half the participants were randomly assigned to receive flashcards describing the classes and their prototypes, while the other half did not receive any such information.
Participants then classified 132 examples and received feedback.
For the Club 1 and Club 2 classes, there were training items with specific structure and frequency:
\begin{itemize}
    \item {\bf 1 Transform, Frequency 1:} Three items differed from their class prototype by one feature and appeared once;
    \item {\bf 2 Transforms, Frequency 1:} Three items differed from their class prototype by two features and appeared once;
    \item {\bf 1 Transform, Frequency 10:} Three items differed from their prototype by one feature and appeared ten times with different distractors;
\end{itemize}
\noindent Additionally, there were three ``Either'' examples that were equally close to the Club 1 and 2 prototypes (1 feature in common with both prototypes) that were seen ten times with different distractors. During training of these instances, participants received positive feedback if they labeled it as either Club 1 or 2 (but not Neither).
Finally, there were thirty examples in the Neither class; if an item had any feature in common with the Neither prototype, then its label was Neither regardless of whether it shared features with the Club prototypes.

During testing, participants were asked to re-classify the Club 1, Club 2, and Either items (they were not tested on the Neither items). 
They were also asked to classify seven novel items not seen during training that included the prototypes of the Club 1, Club 2, Either, and Neither items, as well as three additional Either items.
In total, participants classified 28 test stimuli.
For each test item, participants rated whether the item was old or new (recognition) and their confidence in classifying the item as Club 1 or Club 2 (categorization).
They did not receive feedback on their responses.

\citet{hayesroth1977role} found that reviewing the class prototypes prior to training improved classification performance; however they claimed that the relative ranking of classification accuracy each of the item types was consistent across conditions.
The data also exhibit central tendencies; the prototypes had the highest classification accuracy (even those not seen during training), and the further an item was from its prototypes (the more transforms) the lower its classification accuracy.
Another key finding was that participants were more likely to recognize 1 transformation items that were seen ten times during training than those that were only seen once, but their classification ratings of 1 transformation items was the same regardless of the training frequency.
Next, we aim to show that our theory produces these observed effects.

\begin{figure}[t]
    \centering
    \includegraphics[width=1\linewidth]{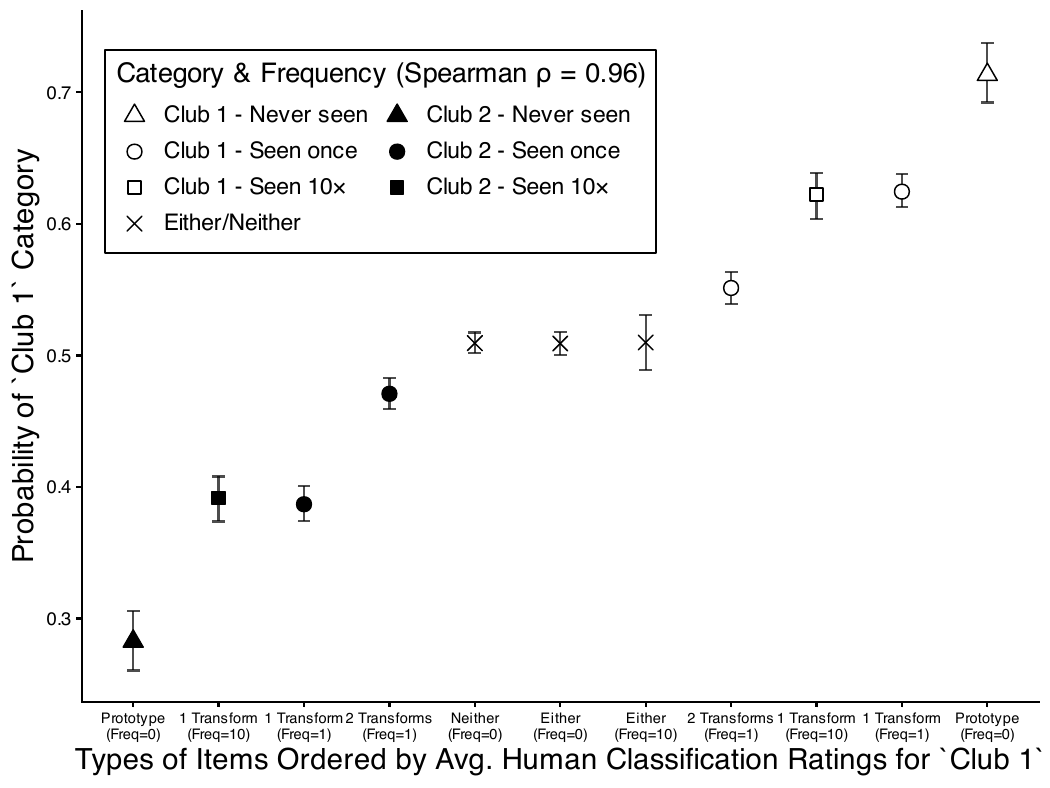}
    \caption{Probabilities of the Club 1 category for the test items in the \citet{hayesroth1977role} experiment, aggregated by item type. Problem types are ordered by avg. human confidence ratings. The legend reports the Spearman rank correlation. Human classification rate data are not available. The error bars represent the 95\% confidence intervals. }
    \label{fig:hayes_roth_classification}
\end{figure}

\subsubsection{Simulation Methodology}

For our evaluation, we simulated the 54 participants in the condition that did not receive information about the class prototypes prior to training.
We focused on this condition because we wanted to demonstrate that our theory exhibits the surprising central tendency result that novel prototypes are more accurately classified than less prototypical items that were previously studied \citep{reed1972pattern,posner1968genesis}.

Our simulation procedure was identical to the one employed for the human experiment.
We trained our Cobweb system by sequentially presenting it with each of the 132 training stimuli along with their associated class labels (Club 1, Club 2, or Neither) in a random ordering.
For Either items, we randomly assigned them to have either a Club 1 or Club 2 label.
Next, we presented the system with the 28 test stimuli (without labels) and generated a predicted probability of the Club 1 class for each item.
We also computed the loglikelihood of each test item given our model, which represents a proxy for the recognition score that participants gave.
To generate these loglikelihood scores, we employed an approach similar to prediction; we expanded nodes with the highest $PMI$ in a best-first fashion up to the max\_nodes limit. We then took the weighted sum the $\log p(x|c)$ values for expanded concepts.
We set $\alpha=1$ and max\_nodes $=100$.

\subsubsection{Results and Discussion}
\begin{figure}[t]
    \centering
    \includegraphics[width=1\linewidth]{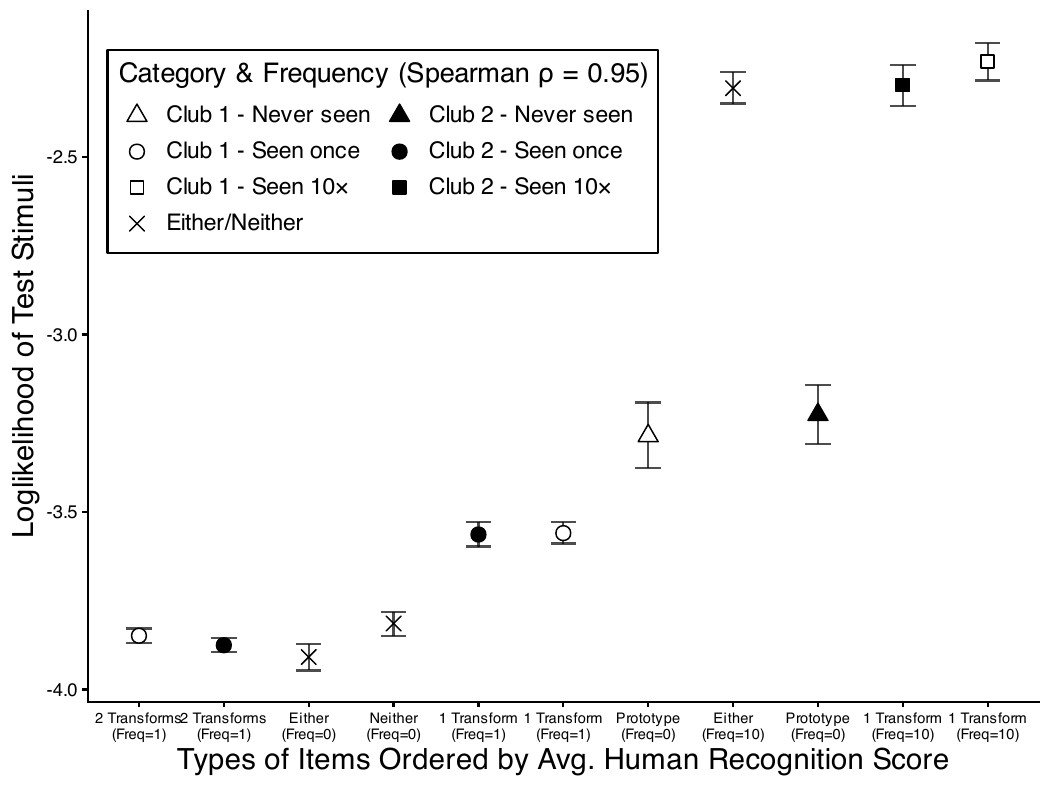}
    \caption{The loglikelihood of test items from the \citet{hayesroth1977role} experiment, aggregated by item type. Problem types are ordered by to avg. human recognition scores.  The legend reports the Spearman rank correlation. The error bars represent the 95\% confidence intervals.}
    \label{fig:hayes_roth_recognition}
\end{figure}

Figures~\ref{fig:hayes_roth_classification} and \ref{fig:hayes_roth_recognition} show the results of our simulations.
It is clear that our theory can account for the central tendencies humans exhibit on this task.
Specifically, Figure~\ref{fig:hayes_roth_classification} shows our system correctly predicts that prototypes will have the highest classification accuracy even though they were not studied during training.
The system also correctly ranks the 1 transform and 2 transform items, with 1 transform items being predicted to have greater accuracy than 2 transform items. 
Lastly, the system predicts the same accuracy for 1 transform items that have been seen once and ten times. This is similar to the human data, which shows a preference for more frequently seen 1 transform items for Club 2 but less frequently seen 1 transform items for Club 1.

Figure~\ref{fig:hayes_roth_recognition} shows that our theory can also largely account for the recognition results reported by \citet{hayesroth1977role}. Mainly, items that have been practiced more frequently have a higher loglikelihood.
Our system ranks the items in a way that closely matches the human recognition scores.
The one exception is the Club 2 prototype, which our system scores as being less recognizable than the humans rated it. 
One possible explanation for this difference is that the recognition scores included data from both experimental conditions,\footnote{\citet{hayesroth1977role} only reported the aggregate classification and recognition ratings, so we cannot look at just those for the condition that did not see the prototypes before training.} including the condition where participants actually saw the prototypes prior to training.
Overall, however, our system's behavior aligns closely with the human results.


\subsection{Non-Linearly and Linearly Separable Categories}
Having shown that our theory can produce a reasonable account of human central tendencies, we  next explored how it compares to other previously established theories of categorization.
Our application of rational analysis was inspired by \citet{anderson1991human}, and the first result he reports is the ability of his theory to account for human data from \citeauthor{medin1978context}'s \citeyearpar{medin1978context} first experiment.
Following \citeauthor{anderson1991human}, we aim to show our theory can also provide an account of these data.

While \citeauthor{anderson1991human} does not report predictions for \citeauthor{medin1978context}'s other experiments, \citet{medin1978context} report predictions from their context model for their second, third, and fourth experiments.
Their context model is an instantiation of the exemplar theory of categorization---it assumes classification decisions are made by comparing a stimulus to stored exemplars, with selective attention weighting on features, and assigning category membership based on overall similarity to category exemplars, rather than abstract concept descriptions (prototypes).
To provide a second basis of comparison, we also simulated their second experiment, which was specifically designed to highlight the shortcomings of independent cue models, which assume that category judgments are based on an additive combination of independently weighted abstract feature cues \citep{franks1971abstraction,hayesroth1977role,reed1972pattern}.
We aim to show that our theory is competitive with these prior theories.

One point worth noting is that our Cobweb instantiation is quite constrained in comparison to these prior approaches. The only adjustable parameters in our system are the max\_nodes, which controls the amount of computation applied at inference time, and the $\alpha$ parameter, corresponding to the strength of the uniform prior across all the discrete features. \citeauthor{anderson1991human}'s model has separate $\alpha$ parameters for each attribute (although usually he sets these to the same value for all attributes),\footnote{Similar to \citeauthor{anderson1991human}, we could have separate $\alpha$ values for each feature, but we prefer as few parameters as possible.} as well as an additional coupling parameter, which controls how likely the system is to form new concepts.
The context and independent cue models, have many more free parameters, such as attentional feature weights.

\begin{table}[t!]
\tiny
\centering
\caption{Training and test stimuli for exp 1. Items have four features: Form (F), Size (S), Color (C), and Position (P).}
\label{tab:medin-exp1-stimuli}
\renewcommand{\arraystretch}{1.2}
\begin{tabular}{|ccccc|ccccc|}
\hline
\multicolumn{10}{|c|}{\textbf{Training Stimuli}} \\
\hline
\multicolumn{5}{|c|}{\textbf{Category A}} &
\multicolumn{5}{c|}{\textbf{Category B}} \\
\hline
Item \# & F & S & C & P & Item \# & F & S & C & P \\
\hline
6 & 1 & 1 & 1 & 1 & 10 & 0 & 0 & 0 & 0 \\
7 & 1 & 0 & 1 & 0 & 15 & 1 & 0 & 1 & 1 \\
9 & 0 & 1 & 0 & 1 & 16 & 0 & 1 & 0 & 0 \\
\hline
\hline
\multicolumn{10}{|c|}{\textbf{Novel Test Stimuli}} \\
\hline
\multicolumn{5}{|c|}{\textbf{Category A}} &
\multicolumn{5}{c|}{\textbf{Category B}} \\
\hline
Item \# & F & S & C & P & Item \# & F & S & C & P \\
\hline
5 & 0 & 1 & 1 & 1 & 3 & 1 & 0 & 0 & 0 \\
13 & 1 & 1 & 0 & 1 & 8 & 0 & 0 & 1 & 0 \\
4 & 1 & 1 & 1 & 0 & 14 & 0 & 0 & 0 & 1 \\
\hline
\end{tabular}
\end{table}

\subsubsection{Human Study Design}

In the first experimental task, thirty-two participants classified stimuli into one of two categories: A or B.
Each example was presented as a geometric forms that varied along four binary dimensions: form, size, color, and position.
During training, participants completed training runs through the 6 training stimuli in Table~\ref{tab:medin-exp1-stimuli}.
For each run, participants were sequentially presented with the six items in a random order, and they classified each example and received feedback.
Participants stopped training when they made no errors on two consecutive runs or when they had completed 20 runs.
After a 5-10 minute distractor task, participants classified all the training stimuli and the novel test stimuli from Table~\ref{tab:medin-exp1-stimuli}.
They classified each example and rated their confidence.
No feedback was given during testing.

The main human effects to predict are: (1) 6 and 10 (the prototypes) should be the most accurate, (2) 10 should have higher accuracy than 6, (3) 15 should be the least accurate among trained (old) items (it is an exception item that shares more features in common with A than B), (4) all the novel A test stimuli in the left column should be more accurate than all novel B test items in the matching row (3, 8, and 14 are highly similar to both an A and B training item, but 5, 13, and 4 are highly similar to two A training items).
The context model correctly predicts all of these effects, but the independent cue model only predicts (1) and (3).

\begin{table}[t!]
\tiny
\centering
\caption{Training and test stimuli for exp 2. Items have four features: Form (F), Size (S), Color (C), and Position (P).}
\label{tab:medin-exp2-stimuli}
\renewcommand{\arraystretch}{1.2}
\begin{tabular}{|ccccc|ccccc|}
\hline
\multicolumn{10}{|c|}{\textbf{Training Stimuli}} \\
\hline
\multicolumn{5}{|c|}{\textbf{Category A}} &
\multicolumn{5}{c|}{\textbf{Category B}} \\
\hline
Item \# & F & S & C & P & Item \# & F & S & C & P \\
\hline
4  & 1 & 1 & 1 & 0 & 12 & 1 & 1 & 0 & 0 \\
7  & 1 & 0 & 1 & 0 & 2  & 0 & 1 & 1 & 0 \\
15 & 1 & 0 & 1 & 1 & 14 & 0 & 0 & 0 & 1 \\
13 & 1 & 1 & 0 & 1 & 10 & 0 & 0 & 0 & 0 \\
5  & 0 & 1 & 1 & 1 &  &  &  &  &  \\
\hline
\hline
\multicolumn{10}{|c|}{\textbf{Novel Test Stimuli}} \\
\hline
\multicolumn{5}{|c|}{\textbf{Category A}} &
\multicolumn{5}{c|}{\textbf{Category B}} \\
\hline
Item \# & F & S & C & P & Item \# & F & S & C & P \\
\hline
1  & 1 & 0 & 0 & 1 & 3  & 1 & 0 & 0 & 0 \\
6  & 1 & 1 & 1 & 1 & 8  & 0 & 0 & 1 & 0 \\
9  & 0 & 1 & 0 & 1 & 16 & 0 & 1 & 0 & 0 \\
11 & 0 & 0 & 1 & 1 &    &  &  &  &  \\
\hline
\end{tabular}
\end{table}

The second experiment followed a similar design to the first.
However, while the first experiment used non-linearly separable items, this experiment used the linearly separable items shown in Figure~\ref{tab:medin-exp2-stimuli}.
The only important procedural difference was that participants stopped training after a single error free run or when they had completed 16 runs.

The experiment was designed such that the context and cue models make opposing predictions for items 7 and 4. 
The context model predicts that 7 should have higher accuracy than 4 because 7 is highly similar to two A patterns and no B patterns, whereas 4 is highly similar to one A pattern and two B patterns.
In contrast, the cue model predicts that 4 should have higher accuracy because it is at least as close as 7 to the prototype no matter how the features are weighted.
The data show that humans are more accurate at predicting 7, which favors the context over the cue model.





\subsubsection{Simulation Methodology}

Our simulations closely mirrored the procedure used in both human experiments.
We ran 32 simulations for each experiment (to match the number of human participants).
During each training run, we generated predicted class labels for each training item and the correctness of these predictions was used to determine whether the stopping conditions had been met.
On average our Cobweb system completed 3 training runs during the first experiment and 2 training runs during the second experiment.
For both experiments, we set max\_nodes $= 100$ and $\alpha = 0.1$.
We tested a few possible values for $\alpha$ and chose the one that seemed to fit the data best.
We found that a value of $0.1$ seemed to more closely fit the data than both $1.0$ and $0.01$.

\begin{figure}
    \centering
    \includegraphics[width=1\linewidth]{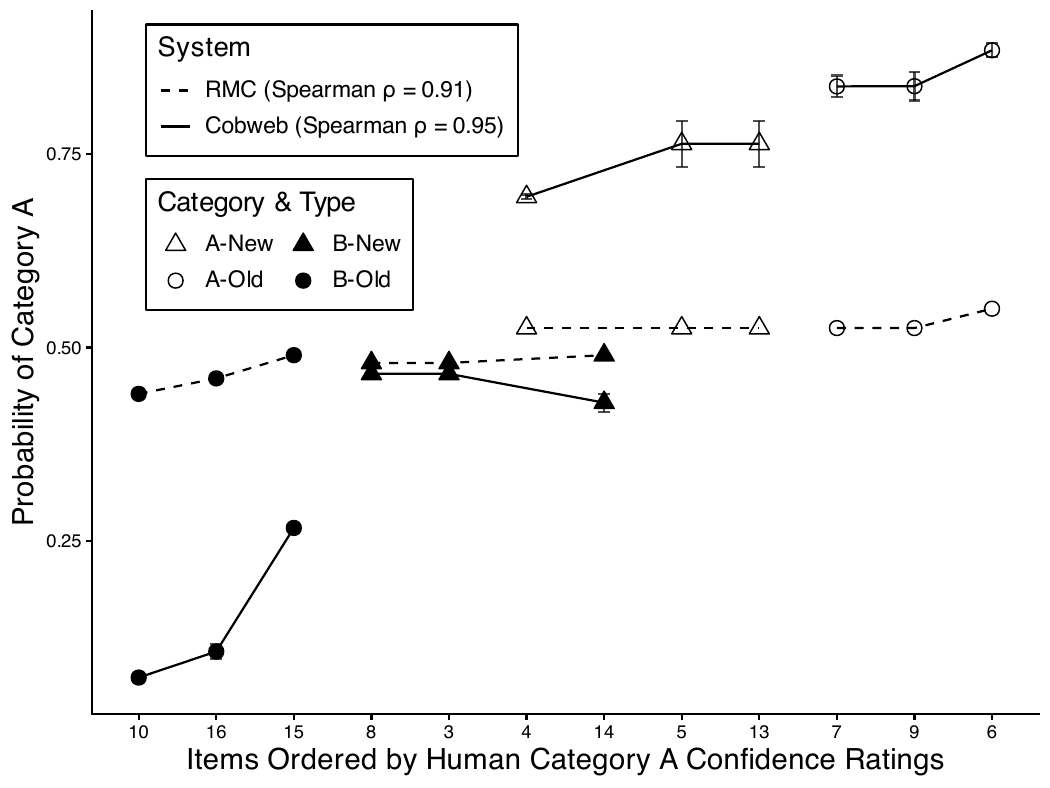}
    \caption{Predicted probability of category A from our system and the RMC for exp 1. The circles denote training items and triangles denote novel test items. The fill denotes item type (A=white, B=black). The lines connect items of the same category and type. Human classification rate data are not available. The error bars denote 95\% confidence intervals.}
    \label{fig:medin_schaffer_exp1}
\end{figure}

We used two approaches to evaluate how well our theory accounts for the data.
First, we looked at whether Cobweb correctly predicts the main effects the experiments were originally designed to test---the ones that differentiate the context and cue models.
Second, we plotted the system's probability of the A category for each test item against the corresponding human confidence ratings.
For comparison purposes, we plotted the predictions from \citeauthor{anderson1991human}'s rational model of categorization (abbreviated as RMC) for experiment 1 and \citet{medin1978context}'s context model for experiment 2.
To evaluate the fit of each, we computed Spearman's rank correlation between the average prediction from each approach and the average human confidence rating for each item.\footnote{We acknowledge that averaging can hide key effects, but we only had access to aggregated scores for humans and comparable systems.}

\begin{figure}
    \centering
    \includegraphics[width=1\linewidth]{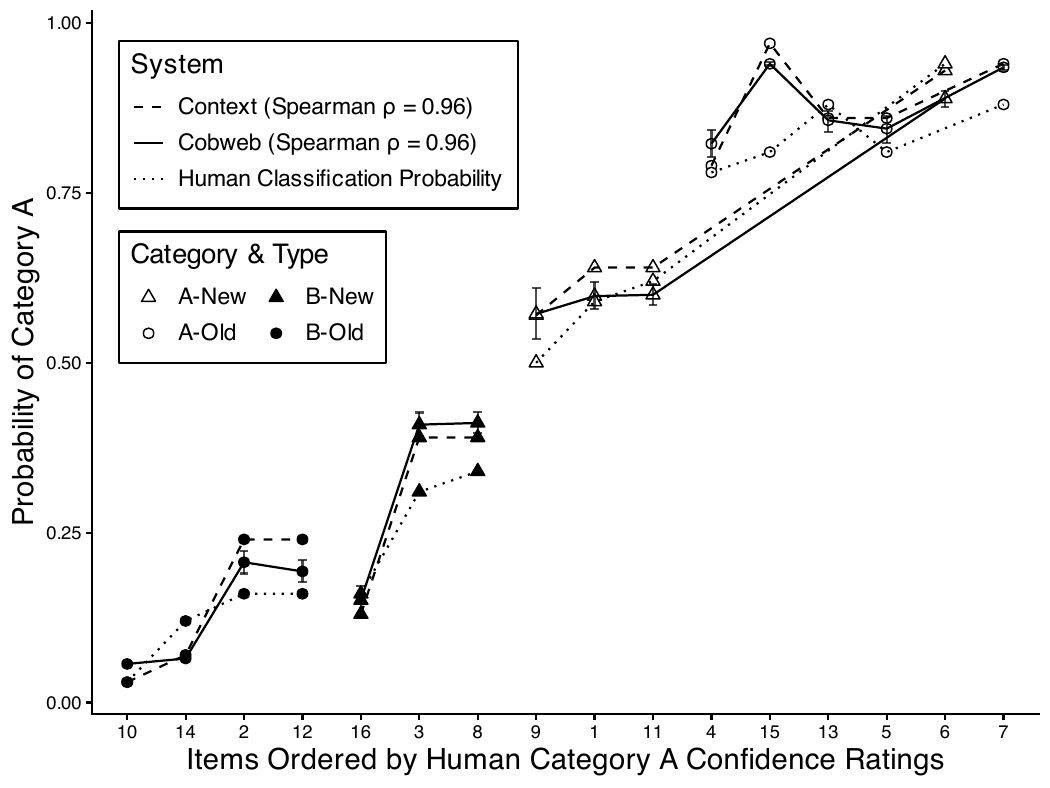}
    \caption{Predicted probability of category A from our system and the context model for exp 2. The circles denote training items, the triangles denote novel test items. The fill denotes the item type (A=white, B=black). The lines connect items of the same category and type. The error bars denote 95\% confidence intervals.}
    \label{fig:medin_schaffer_exp2}
\end{figure}

\subsubsection{Results and Discussion}

\begin{figure*}[ht!]
    \centering
    \includegraphics[width=0.75\linewidth]{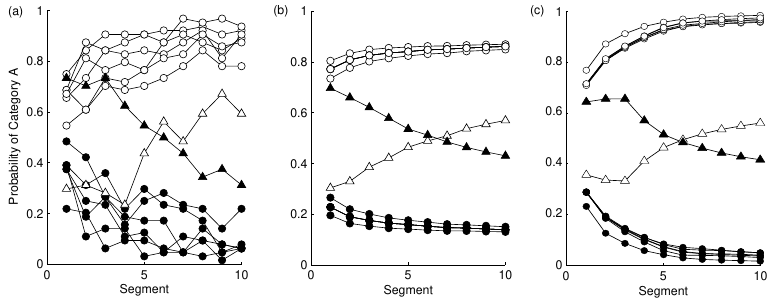}
    \caption{Predictions and human data for \citet{smith1998prototypes}'s second experiment. Each line is a stimuli with Category A denoted by white, Category B denoted by black, and exception items denoted by triangles. Figures (a) and (b) are reproduced from \citet{griffiths2007unifying}, showing the human and HDP model data respectively. Our system's predictions are shown in (c).
    }
    \label{fig:exemplar-prototype}
\end{figure*}

Figure~\ref{fig:medin_schaffer_exp1} and \ref{fig:medin_schaffer_exp2} show our results.
Increasing predicted probabilities from left to right in both figures is indicative of high agreement with the human data.
In experiment 1, our Cobweb system better matches the data better than the RMC (spearman correlation of 0.95 vs. 0.91).\footnote{The RMC predictions look different but are ordered similarly and thus achieve a similar rank correlation.}
\citet{anderson1991human} reports that his model produces fits ``as good as those obtained by \citeauthor{medin1978context}''.
Our system also fits the experiment 2 data well, producing nearly identical predictions to the context model.
Furthermore, it successfully predicts all of the main effects that the two experiments were designed to test (effects 1-4 for experiment 1 and that item 7 will have higher accuracy than item 4 for experiment 2).
Our theory can account for these human categorization data as well as (or better) than the RMC, context, and cue models.

\subsection{Prototype-Exemplar Transition}
A longstanding debate concerns whether categorization is better explained by prototype or exemplar theories.
Rather than aligning strictly with either theory, \citet{smith1998prototypes} argue that people exhibit behavior more consistent with prototype theory in some situations and exemplar theory in others.
Our theory has the potential to exhibit both types of behavior because the leaves of Cobweb's categorization tree correspond to exemplars, whereas the intermediate nodes correspond to abstract prototypes.\footnote{Although Cobweb's concepts can be viewed as abstract prototypes, they need not align with provided category labels.}
To explore this potential, we simulated the non-linearly separable condition of \citeauthor{smith1998prototypes}'s \citeyearpar{smith1998prototypes} second experiment.
This also gave us an opportunity to compare to the Hierarchical Dirichlet Process (HDP) approach proposed by \citet{griffiths2007unifying}, which they present as an extension of \citeauthor{anderson1991human}'s RMC.

\subsubsection{Human Study Design}
For this task, 16 participants were trained to categorize 14 six-letter nonsense words belonging to one of two categories. 
The items were designed such that each letter slot could take on only one of two values (six boolean features).
Within each category, there were two ``prototype'' items that shared no features in common (000000 and 111111), five ``standard'' items that shared five features with their respective prototype (e.g., 100000, 010000, 001000, 000100, and 000010 for the 000000 prototype), and one ``exception'' item that only had one feature in common with the prototype (e.g., 111101 for the 000000 prototype).
A key characteristic of the categories is that they are not linearly separable.
Participants were presented with randomly shuffled blocks of the 14 stimuli and asked to categorize each stimuli, receiving feedback after each decision.
They completed 40 blocks (560 categorization trials) and the responses were grouped into 10 segments that each contained four blocks. 

\citet{griffiths2007unifying} reports participants' average predictions on each item across the segments (see Figure~\ref{fig:exemplar-prototype}a)
Their analysis shows that on average participants correctly classify the prototypes and standard items initially and that their accuracy improves with practice.
In contrast, participants miscategorize the exception items initially.
They show no change in accuracy over the first 3 segments, but their accuracy starts improving around segment 4 and keeps improving over the remaining segments.
About half way, participants begin correctly categorizing the exception items on average.
Both \citet{smith1998prototypes} and \citet{griffiths2007unifying} claim that the data is more consistent with prototype theory in early segments, but more consistent with exemplar theory in later segments.
We aim to show that our theory predicts this effect.

\subsubsection{Simulation Methodology}
We trained our system following the experimental procedure.
In alignment with \citet{griffiths2007unifying}, we generated predictions for all 14 stimuli after each training segment.
We conducted 16 simulations, corresponding to the 16 participants.
We set $\alpha=2.0$ and max\_nodes $=100$.
To check for the transition effect, we plotted the average predicted probability of category A for each item after each segment.
We reproduced plots from \citet{griffiths2007unifying} showing the human data and HDP predictions.\footnote{Unlike \citet{griffiths2007unifying} and \citet{smith1998prototypes}, we did not incorporate guessing parameters (to capture the possibility of random participant responses) into our system.}

\subsubsection{Results and Discussion}
Figure~\ref{fig:exemplar-prototype} shows the average probability of category A for each stimuli over the 10 training segments.
Our system very clearly exhibits the prototype-exemplar transition, where the exception items are misclassified early in learning, show little change in accuracy over the first few segments, but are correctly classified later in training.

The aim here was simply to demonstrate that our theory could {\it qualitatively} account for the prototype-exemplar transition.
Therefore, we arbitrarily set the values of $\alpha$ and max\_nodes to reasonable values that seemed to exhibit the effect.
We tested several possible values for $\alpha$ and found that it seemed to shift where the exception items transitioned from incorrect to correct and that it was possible to choose values where the transition happened during the first segment or after the final segment. For example, when $\alpha < 0.6$, the system correctly classifies all items (including the exception items) starting at the first segment.
When $\alpha=1$ our results look similar to the HDP predictions.
However, we discovered that when we set $\alpha=2$, we our system exhibits the phase change behavior observed in Figure~\ref{fig:exemplar-prototype}c, where no improvement in the exception items is observed in the first few segments.
We believe that the greater value for $\alpha$ leads the system to favor the formation and use of prototypes as it needs to form larger categories to overpower the stronger uniform feature prior.
The phase change occurs when the system gains enough exposures to the exception that it forms distinct concepts (i.e., prototypes) to represent them. 




\section{Conclusion and Future Work}

We conduct a rational analysis of human categorization grounded in information theory.
Unlike the prior rational analysis by \citet{anderson1991human}, we assume people learn taxonomically organized categories that have maximal mutual information with the features of their experiences.
These assumptions yield a new theory that we instantiated within a new version of Cobweb.
Using our variant, we simulated several classic categorization experiments, and found that our system can predict the human behavior as well as (or better) than the independent cue and context models, the rational model of categorization, and the hierarchical Dirichlet process model.
These results provide promising initial evidence to support our theory.
In the future, we plan to run ablations to test the importance of the taxonomic and mutual information assumptions individually and to run experiments to distinguish where our theory's predictions differ from those of alternatives.
Lastly, we hope to investigate if it can account for a wider range of effects, with the goal of producing a new unified theory.






\section{Acknowledgments}
This project is partially supported by AFOSR Grant No. FA9550-23-1-058.
Any opinions, findings, and conclusions or recommendations expressed in this material are those of the author(s) and do not necessarily reflect the views of the funding agency.


\printbibliography

@article{smith1998prototypes,
  title={Prototypes in the mist: The early epochs of category learning.},
  author={Smith, J David and Minda, John Paul},
  journal={Journal of Experimental Psychology: Learning, memory, and cognition},
  volume={24},
  number={6},
  pages={1411},
  year={1998},
  publisher={American Psychological Association}
}

@article{fisher1990structure,
  title={The structure and formation of natural categories},
  author={Fisher, Douglas and Langley, Pat},
  journal={Psychology of Learning and Motivation},
  volume={26},
  pages={241--284},
  year={1990},
  publisher={Elsevier}
}

@article{posner1968genesis,
  title={On the genesis of abstract ideas.},
  author={Posner, Michael I and Keele, Steven W},
  journal={Journal of experimental psychology},
  volume={77},
  number={3p1},
  pages={353},
  year={1968},
  publisher={American Psychological Association}
}

@article{howes2009rational,
  title={Rational adaptation under task and processing constraints: implications for testing theories of cognition and action.},
  author={Howes, Andrew and Lewis, Richard L and Vera, Alonso},
  journal={Psychological review},
  volume={116},
  number={4},
  pages={717},
  year={2009},
  publisher={American Psychological Association}
}

@article{church1990word,
  title={Word association norms, mutual information, and lexicography},
  author={Church, Kenneth and Hanks, Patrick},
  journal={Computational linguistics},
  volume={16},
  number={1},
  pages={22--29},
  year={1990}
}

@article{fano1961transmission,
  title={Transmission of information: A statistical theory of communications},
  author={Fano, Robert M and Hawkins, David},
  journal={American Journal of Physics},
  volume={29},
  number={11},
  pages={793--794},
  year={1961},
  publisher={AIP Publishing}
}

@article{corter1992explaining,
  title={Explaining basic categories: Feature predictability and information.},
  author={Corter, James E and Gluck, Mark A},
  journal={Psychological bulletin},
  volume={111},
  number={2},
  pages={291},
  year={1992},
  publisher={American Psychological Association}
}

@article{fisher1987knowledge,
  title={Knowledge acquisition via incremental conceptual clustering},
  author={Fisher, Douglas},
  journal={Machine learning},
  volume={2},
  number={2},
  pages={139--172},
  year={1987},
  publisher={Springer}
}

@article{reed1972pattern,
  title={Pattern recognition and categorization},
  author={Reed, Stephen K},
  journal={Cognitive psychology},
  volume={3},
  number={3},
  pages={382--407},
  year={1972},
  publisher={Elsevier}
}

@article{franks1971abstraction,
  title={Abstraction of visual patterns.},
  author={Franks, Jeffery J and Bransford, John D},
  journal={Journal of experimental psychology},
  volume={90},
  number={1},
  pages={65},
  year={1971},
  publisher={American Psychological Association}
}

@article{barari2024cobweb4v,
title = {Robust incremental learning of visual concepts without catastrophic forgetting},
journal = {Cognitive Systems Research},
volume = {96},
pages = {101447},
year = {2026},
issn = {1389-0417},
doi = {https://doi.org/10.1016/j.cogsys.2026.101447},
url = {https://www.sciencedirect.com/science/article/pii/S1389041726000148},
author = {Nicki Barari and Xin Lian and Christopher J. MacLellan}
}

@inproceedings{lian2024cobweb,
  title={Cobweb: An incremental and hierarchical model of human-like category learning},
  author={Lian, Xin and Varma, Sashank, and MacLellan, C. J.},
  booktitle={Proceedings of the Annual Meeting of the Cognitive Science Society},
  year={2024}
}

@inproceedings{griffiths2007unifying,
  title={Unifying rational models of categorization via the hierarchical Dirichlet process},
  author={Griffiths, Thomas L and Canini, Kevin R and Sanborn, Adam N and Navarro, Daniel J},
  booktitle={Proceedings of the Annual Meeting of the Cognitive Science Society},
  volume={29},
  number={29},
  year={2007}
}

@article{anderson1991human,
  author = {Anderson, John R.},
  title = {The adaptive nature of human categorization},
  journal = {Psychological Review},
  year = {1991},
  volume = {98},
  number = {3},
  pages = {409--429}
}

@article{hayesroth1977role,
title = {Concept learning and the recognition and classification of exemplars},
journal = {Journal of Verbal Learning and Verbal Behavior},
volume = {16},
number = {3},
pages = {321-338},
year = {1977},
issn = {0022-5371},
doi = {https://doi.org/10.1016/S0022-5371(77)80054-6},
url = {https://www.sciencedirect.com/science/article/pii/S0022537177800546},
author = {Barbara Hayes-Roth and Frederick Hayes-Roth}
}

@article{medin1978context,
  author = {Medin, Douglas L. and Schaffer, Marguerite M.},
  title = {Context theory of classification learning},
  journal = {Psychological Review},
  year = {1978},
  volume = {85},
  number = {3},
  pages = {207--238}
}

\end{document}